\documentclass[preprint]{article}
\usepackage[]{neurips_2020}
  
\usepackage[utf8]{inputenc} 
\usepackage[T1]{fontenc}    
\usepackage{hyperref}       
\usepackage{url}            

\usepackage{booktabs}       
\usepackage{amsfonts}       
\usepackage{nicefrac}       
\usepackage{microtype}      

\usepackage{wrapfig}

\usepackage[mathcal]{euscript}
\usepackage{pstricks,comment}
\usepackage{mathtools}
\usepackage{units}
\usepackage{soul}
\usepackage{ulem}

\usepackage{dutchcal}

\usepackage{xparse}

\usepackage{accents}

\allowdisplaybreaks

\newcommand{\bm}[1]{\mbox{\boldmath{$#1$}}}

\newcommand{\bx}{\bm{x}}

\newcommand{\bW}{\bm{W}}

\newcommand{\bh}{\bm{h}}

\newcommand{\bS}{\bm{S}}

\newcommand{\RR}{\mathbb{R}}

\newcommand{\tensor}[1]{\bm{\cal #1}}

\DeclarePairedDelimiterX{\norm}[1]{\lVert}{\rVert}{#1}

\newcommand{\order}[1]{{\cal O}(#1)}
\newcommand{\Order}[1]{{\cal O}\left(#1\right)}

\newtheorem{corollary}{Corollary}
\newtheorem{theorem}{Theorem}

\newcommand{\qedsymbol}{\rule{2mm}{2mm}}

\title{Tensor train decompositions on recurrent networks}

\author{%
  Alejandro Murua\\
  D\'epartement de math\'ematiques et de statistique\\
  Universit\'e de Montr\'eal\\
  2920, chemin de la Tour, H3T 1J4\\
    Montreal, Qu\'ebec, Canada\\
    \texttt{murua@dms.umontreal.ca}\\
    \And
    Ramchalam Ramakrishnan\\ 
    Huawei Noah's Ark Research Lab\\
    Montreal Research Center\\
    7101 Avenue du Parc, H3N 1X9\\
    Montreal Qu\'ebec, Canada\\
    \texttt{ramchalam.ramakrishnan@huawei.com}\\
    \And
    Xinlin Li\\
    Huawei Noah's Ark Research Lab\\
    Montreal Research Center\\
    7101 Avenue du Parc, H3N 1X9\\
    Montreal Qu\'ebec, Canada\\
    \texttt{xinlin.li1@huawei.com}
    \And
    Rui Heng Yang\\
    Huawei Noah's Ark Research Lab\\
    Montreal Research Center\\
    7101 Avenue du Parc, H3N 1X9\\
    Montreal Qu\'ebec, Canada\\
    \texttt{rui.heng.yang@huawei.com}
    \And    
    Vahid Partovi Nia\\
    Huawei Noah's Ark Research Lab\\
    Montreal Research Center\\
    7101 Avenue du Parc, H3N 1X9\\
    Montreal Qu\'ebec, Canada\\    
    \texttt{vahid.partovinia@huawei.com}
}

\begin{document}
\maketitle

\begin{abstract}
  Recurrent neural networks (\textsc{rnn}) such as long-short-term memory (\textsc{lstm}) networks are essential in a multitude of daily live tasks such as speech, language, video, and multimodal learning. The shift from cloud to edge computation intensifies the need to contain the growth of \textsc{rnn} parameters. Current research on \textsc{rnn} shows that despite the performance obtained on convolutional neural networks (\textsc{cnn}), keeping a good performance in compressed \textsc{rnn}s is still a challenge. Most of the literature on compression focuses on \textsc{cnn}s using matrix product (\textsc{mpo}) operator tensor trains. However, matrix product state (\textsc{mps}) tensor trains have more attractive features than \textsc{mpo}s, in terms of storage reduction and computing time at inference. We show that \textsc{mps} tensor trains should be at the forefront of \textsc{lstm} network compression  through a theoretical analysis and practical experiments on \textsc{nlp} tasks.
\end{abstract}

Since the beginning of the last decade, computational resources have
taken a primary role in the advancing of artificial intelligence. With the advent of
new architectures for neural networks, starting with AlexNet and convolutional neural networks,
computational demand has accelerated to the point of doubling itself each hundred days \citep{openai.blog:November-2019}. In fact from AlexNet to AlexZero, computing demand has increased 3000 times.
This implies major challenges for resources such as memory. Neural network model size is
a bottleneck. The use of memory and computer power for training neural nets is enormous.
For example, AlexNet
takes 1.5 weeks to train on ImageNet on one Nvidia Titan\textsc{x} \textsc{gpu} \citep{Hansetal-2015}.
Although this might seems like a long time, we have to consider that following the current
trend in neural network modeling, a more sophisticated architecture, such as ResNet152, might take
at least ten times longer to train.
Model size is indeed a major challenge:
larger model implies larger sizes, which implies higher energy consumption.
 In a recent tweet debate, Elliot Turner, the \textsc{ceo} and co-founder of Hologram \textsc{ai},
wrote that it costs \$ 245,000 and 2.5 days on 512 \textsc{tpu} v3 chips for 500K steps with an Adam optimizer, to train the \textsc{xln}et model on natural language tasks \citep{syncedreview.com:June-2019}.
 General purpose microprocessors are not getting faster or more efficient. 
This in turn, has lead to specialized domain specific hardware for improvements in inference speed and accuracy, such as neural processing units (\textsc{npu}), and tensor processing units (\textsc{tpu}, and Edge \textsc{tpu}).

The question is how to capitalize on such breakthrough models so as to overcome these challenges.
We believe part of the answer consists on performing model compression and/or quantization, or pruning.
The idea is to enable the deployment of state-of-the-art models in severely resource constrained environments.
Running deep learning models on low-resource devices is a highly-promising area of current research in artificial intelligence (AI).
This is sometimes referred to as Tiny-AI \citep{tinyML}. \textsc{mit} Technology Review has identified it as one of the top 10 promising technologies of 2020
\citep{MIT-2020}.
The present work is primarily about storage compression and compression for compute-constraint efficient inference
on recurrent neural networks (\textsc{rnn}) and
long-short-term memory networks (\textsc{lstm}) based architectures.
We show that \textsc{lstm} architectures based on weight matrices given by tensor trains can achieve performances similar or better
than classical \textsc{lstm} models, even when the number of parameters is reduced by half or even 80\% with respect a
classical {\it full} \textsc{lstm} model. Furthermore, we show theoretically and in practice through a series of
experiments, that some tensor train networks can do inference in a fraction of time.

\textsc{lstm} networks are
essential in automatic speech recognition models, language modeling, machine translation, and handwritten recognition among many other difficult tasks.
They have a simple but heavy architecture that requires the estimation of eight weight matrices during training,
and the same number of matrix-vector multiplications during both inference and training.
Therefore, the deployment of large \textsc{lstm} based models on edge devices consumes substantial storage, memory and computational resources.
Because of the dependencies of time steps, the operations in \textsc{lstm} networks are not easy to parallelize. Although, specialized hardware such as NPU, and TPU (specifically for edge devices)
support sigmoid and hyperbolic tangent operations, these hardware are not optimized for a large number of such operations.

Compression in the form of tensor decomposition or low-rank matrix approximations is about giving sound structure to the gate weights, so as to reduce the number of parameters \citep{pan-et-al-2019,tjandra-et-al-2018,tjandra-et-al-2017,kusupati-2019,winata-et-al-2018,winata-et-al-2019,barone-2016}.
Alternatives to compression are quantization \citep{li-et-al-2018,alom-et-al-2017,he-et-al-2016,hubara-et-al-2016}
  and pruning \citep{ramakrishnan-et-al-2019,luo2017thinet,Han2016DeepCC,hu-et-al-2016}.
Quantization is about reducing the computational time of multiplications of matrices and vectors
using few precision-bits. Although this reduces memory use, it does not reduce the number of parameters.
Pruning is about eliminating unnecessary weights (elements of the weight matrix) or even nodes in the network. However, pruning trains with full precision and hence it does not reduces
memory during training, nor precision bits.
Moreover, pruning patterns in the weight matrix might be chaotic, which implies non-standard ways to store information.

Even though several studies have shown that compression and quantization
do not hurt performance with feedforward neural networks
nor convolution neural networks (\textsc{cnn}),
research with recurrent
neural networks has shown that keeping
a good level of precision in the computations is very challenging.
However, we show that when adequate tensorization techniques are applied, 
the performance of compressed \textsc{lstm} networks on natural language tasks is not affected.
We shed light onto how two simple tensorization techniques are able to achieve even better performance
than a full network model. These are
the so-called matrix product operator (\textsc{mpo}) and the matrix product state (\textsc{mps}) tensor trains.
We start with a brief introduction of these tensor decompositions and its compression properties.
In Section~\ref{sec:storage:time}, we show how in theory \textsc{mps} can be used efficiently for inference.
Section~\ref{sec:initialization} deals with training issues arising from tensor-train architectures.
In Section~\ref{sec:KD}, we shed light on how to improve the performance of tensor train architectures
using knowledge distillation and regularization.
Our theoretical results are corroborated in Section~\ref{sec:experiments} through 
an extensive study on the performance of tensor train-based \textsc{lstm} architectures on a \textsc{nlp} task.

\section{Tensorization and Compression Techniques}
\label{sec:compression}
A tensor is usually denoted by calligraphic bold symbols, e.g. $\tensor{W}$.
For compression of weight matrices, we could embed the matrix into a tensor as follows.
Suppose that we can write $N= I_1\times I_2 \times \cdots \times I_n$, and
$M= J_1\times J_2 \times \cdots \times J_m$, for some positive integers $N, M, n,m$.
Then we can think of the matrix $\bW$ as a tensor $\tensor{W} \in \mathbb{R}^{ I_1\times I_2 \times \cdots \times I_n
  \times  J_1\times J_2 \times \cdots \times J_m}$. 
The element $w_{ij}$ of the matrix $\bW$ is mapped to the element
$\tensor{W}(i_1,i_2,\cdots , i_n,j_1,j_2,\cdots , j_m)$ of the tensor,
so that, for example, following the big-endian colexicographic ordering, we map
$i = i_n + (i_{n-1}-1)I_n  + (i_{n-2}-1)I_{n-1} I_n + \cdots + (i_1 - 1) I_2 \cdots I_n$, and
$j = j_m + (j_{m-1}-1)J_m  + (j_{m-2}-1)J_{m-1} J_m + \cdots + (j_1 - 1) J_2 \cdots J_m$.
For the sake of space, in what follows we briefly described
    two tensor train decompositions used recently for compressing neural networks.
Other tensor decompositions also used in compression of \textsc{cnn}s are the Polyadic and Tucker decompositions \citep{kolda-bader-2009},
and the tensor ring decomposition \cite{pan-et-al-2019}, also known as tensor-chain in the Physics literature \citep{cichocki-et-al-2016}.
    
\begin{figure}[ht!]
  \centering
  \includegraphics[width=0.90\linewidth]{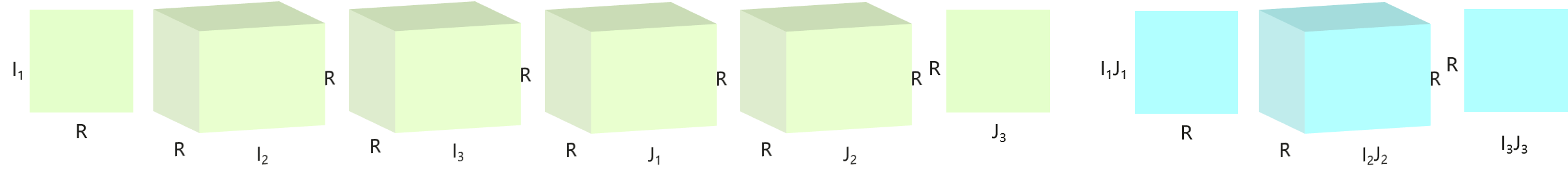} 
\caption{A six-core \textsc{mps} (left panel) and a three-core \textsc{mpo} (right panel). R represents the inner ranks of the cores.}
\label{fig:MPS3}
\end{figure}

\paragraph{Tensor-train decomposition.}
The tensor-train decomposition, also known as the matrix product state (\textsc{mps}) in Physics, decomposed
the tensor $\tensor{W} \in \RR^{I_1 \times \cdots \times I_n \times J_1 \times \cdots \times J_m}$
 into a product of $n+m$ core tensors
$\tensor{W} = \tensor{A}^{(1)} \times^1 \tensor{A}^{(2)} \times^1 \cdots \times^1 \tensor{A}^{(n)}
\times^1 \tensor{B}^{(1)} \times^1 \cdots \times^1 \tensor{B}^{(m)},$
so that 
$\tensor{A}^{(k)} \in \RR^{r_{r(k-1)} \times I_k \times r_{rk}}$,
$\tensor{B}^{(k)} \in \RR^{r_{c(k-1)} \times J_k \times r_{ck}}$,
$r_{r0} = r_{cm} = 1$, $r_{rn}=r_{c0}$.
We refer to the set of indices $\{ r_{rk}\},$ and $\{ r_{ck}\}$ as the {\it inner ranks} of the core tensors. 
Here the notation $\tensor{G}^{(k)} \times^1 \tensor{G}^{(k+1)}$ stands for the
mode-$(3,1)$ product between two tensors \citep{cichocki-et-al-2016}.
The leftmost panel of Figure~\ref{fig:MPS3} depicts an \textsc{mps} tensor train with six cores.

\noindent {\bf The matrix product operator (\textsc{mpo}) tensor.}
A special case of tensor train corresponds to the case 
$\tensor{W} \in  \RR^{ I_1\times I_2 \times \cdots \times I_n   \times  J_1\times J_2 \times \cdots \times J_n}$,
that is, when $m=n$. In this case the tensor may be rewritten as
$\tensor{W}= \tensor{A}^{(1)} \times^1 \tensor{A}^{(2)}\times^1 \cdots \times^1\tensor{A}^{(n)}$
with $\tensor{A}^{(k)} \in  \RR^{ r_{k-1} \times I_kJ_k \times r_k}$, $k=1,\ldots,n$, with
$r_0=r_n = 1$.
The indexes are organized so that in the notation
$\tensor{W}( h_1, h_2, \ldots, h_n)$, 
$h_k$ corresponds to $(i_k, j_k)$ in the form $i_k + (j_k-1) I_k$.
$i_k \in\{1, \ldots, I_k\}$, $j_k\in \{1, \ldots, J_k\}$, $k=1,\ldots, n$.
In an \textsc{mpo} the input and output entries are intertwined.
The right panel of Figure~\ref{fig:MPS3} depicts an \textsc{mpo} tensor train with three cores.
Note that a two-core \textsc{mpo} is a two-matrix decomposition often more efficient than a
reduced matrix decomposition (e.g., singular value decomposition) of the weight matrix.

\section{Storage and Computing time}
\label{sec:storage:time}

In addition to the compression properties of \textsc{mps} and \textsc{mpo} tensor train decompositions,
we are interested
in the order of operations required at inference, that is, to obtain the products
$\bW \bx$ between the weight matrices and the input or output vectors involved in the computations
of the gates of a \textsc{lstm} unit.

Let  $R= \max\{ r_{c0}, \ldots, r_{cm}, r_{r0}, \ldots, r_{rn}\}$,
and let  $I= \max\{I_1, \ldots, I_n\}$, and $J=\max\{ J_1, \ldots, J_m\}$.
Also, denote by $O_{\textsc{mps}}$ and $O_{\textsc{mpo}}$, respectively, the order of the number of operations required at inference by
an \textsc{mps} and an \textsc{mpo} \textsc{lstm} cell.
The main result of this section is the following.
\begin{theorem}\label{theo:1}
   The storage required for an \textsc{mps} tensor train  is of the order $\Order{R(I+J) + R^2[(n-1)I +(m-1)J]}$,
   while for an \textsc{mpo} tensor train is $\Order{I J[ 2R + (n-2) R^2]}$.
   Furthermore, the number of operations required to compute $\bW\bx$ for \textsc{mps} is order $O_{\textsc{mps}} = \Order{R (N+M) + R^2[(n-1)N +(m-1)M]}$.
 For \textsc{mpo}, the weight matrix needs to be reconstructed, so the multiplication requires $O_{\textsc{mpo}} = \Order{N M[R + R^2(n-2)]}$.
 \end{theorem}

\paragraph{Proof:}
For \textsc{mps}, the dimensions of the core tensors are
$(1\times I_1\times r_{r1}, r_{r1} \times I_2 \times r_{r3}, \ldots, r_{r(n-1)} \times I_n \times
r_{c0}, r_{c0} \times J_1\times r_{c1}, \ldots, r_{c (m-1)} \times J_m \times 1)$.
The storage needed for this decomposition is
$\sum_{i=1}^n r_{r(i-1)} r_{ri} I_i + \sum_{j=1}^m r_{c(j-1)} r_{cj} J_j,$
where $r_{r0}= r_{cm} = 1$. Thus, we have the upper bound
$R(I+J) + R^2[(n-1)I + (m-1)J].$
For \textsc{mpo}, the storage needed is
$\sum_{i=1}^n r_{i-1} r_{i} I_iJ_i,$
where $r_{0}= r_{n} = 1$, which is bounded from above by
$2RIJ + R^2 (n-2)IJ$.

Next, turning to the computation aspects of tensor train, 
let us denote by $ \underaccent{\tilde}{j} = (j_1,\ldots, j_m)$, $\underaccent{\tilde}{i} = (i_1,\ldots, i_n)$,
$\underaccent{\tilde}{k} = (k_1,\ldots, k_{n-1})$, $\underaccent{\tilde}{h} = (h_1,\ldots, h_{m-1})$.
For \textsc{mps}, by organizing the terms adequately, one can compute $ \tensor{W}(\underaccent{\tilde}{i})$ as
    {\small
\begin{multline*}
     \tensor{W}(\underaccent{\tilde}{i}) = \sum_{\underaccent{\tilde}{j}}
      \sum_{\underaccent{\tilde}{k}}
            \sum_{h_o =1}^{r_{c0}} \sum_{\underaccent{\tilde}{h}}
\tensor{A}^{(1)}_{(1, i_1, k_1)} \cdots
\tensor{A}^{(n)}_{(k_{n-1}, i_n, h_o)} 
\tensor{B}^{(1)}_{(h_o, j_1, h_1)} \cdots
\tensor{B}^{(m)}_{(h_{m-1}, j_m, 1)}
\tensor{X}(\underaccent{\tilde}{j}) \\
= \sum_{\underaccent{\tilde}{h}}
 \sum_{h_o =1}^{r_{c0}}  \biggl(
\sum_{\underaccent{\tilde}{k}}
\tensor{A}^{(1)}_{(1, i_1, k_1)}  \cdots
\tensor{A}^{(n)}_{(k_{n-1}, i_n, h_o)} \biggr) 
\sum_{\underaccent{\tilde}{j}}
\tensor{B}^{(1)}_{(h_o, j_1, h_1)}  \cdots
\tensor{B}^{(m)}_{(h_{m-1}, j_m, 1)}
\tensor{X}(\underaccent{\tilde}{j}) \\
= \sum_{h_o =1}^{r_{c0}}
\tensor{F}_{h_o}(\underaccent{\tilde}{i})
\sum_{\underaccent{\tilde}{j}}\tensor{X}(\underaccent{\tilde}{j}) \sum_{\underaccent{\tilde}{h}}
\tensor{B}^{(1)}_{(h_o, j_1, h_1)}  \cdots
\tensor{B}^{(m)}_{(h_{m-1}, j_m, 1)}
= \sum_{h_o =1}^{r_{c0}}
\tensor{F}_{ h_o}(\underaccent{\tilde}{i})
\sum_{\underaccent{\tilde}{j}}\tensor{X}(\underaccent{\tilde}{j}) 
\tensor{G}_{h_o}(\underaccent{\tilde}{j}),
    \end{multline*}
with $\tensor{F}_{h_o}(\underaccent{\tilde}{i})= \sum_{\underaccent{\tilde}{k}}
\tensor{A}^{(1)}_{(1, i_1, k_1)}  \cdots \tensor{A}^{(n)}_{(k_{n-1}, i_n, h_o)}$, and
$\tensor{G}_{h_o}(\underaccent{\tilde}{j}) = \sum_{\underaccent{\tilde}{h}}
\tensor{B}^{(1)}_{(h_o, j_1, h_1)}  \cdots \tensor{B}^{(m)}_{(h_{m-1}, j_m, 1)}$.
    The tensors $\tensor{F}_{h_o}$ and $\tensor{G}_{h_o}$ can be pre-computed.
    The number of operations requires is of order
    $\order{r_{c0} [N+M]}$.
    The sums for all $h_{m-2},$ 
    $\tensor{D}_{h_{m-1}}(j_{m-1},j_m) = \sum_{h_{m-1}} \tensor{B}^{(m-1)}_{(h_{m-2}, j_{m-1}, h_{m-1})}
    \tensor{B}^{(m)}_{(h_{m-1}, j_m, 1)},$
    can be computed in $\order{r_{c(m-2)} r_{c(m-1)} }$.
    Hence, the sums for all $h_{m-3}$, 
    $\tensor{D}_{h_{m-2}}(j_{m-2},j_{m-1}, j_m) = \sum_{h_{m-2}} \tensor{B}^{(m-2)}_{(h_{m-3}, j_{m-2}, h_{m-2})}
    \tensor{D}_{h_{m-1}}(j_{m-1}, j_m)$
    can be computed in $\order{r_{c(m-3)}r_{c(m-2)} + r_{c(m-2)} r_{c(m-1)}}$.
    Continuing with this reasoning, we conclude that 
    the sums,  for all $h_o$, $\tensor{G}_{h_o}(\underaccent{\tilde}{j})$
    can be computed in $\order{\sum_{j=1}^{m-1} r_{c(j-1)} r_{cj}}$.
    Similarly, the sums $\tensor{F}_{h_o}(\underaccent{\tilde}{i})$ can be computed  in order
    $\Order{ \sum_{i=1}^{n-1} r_{r i}r_{r(i+1)}}$, with $r_{rn} = r_{c0}$.
    Thus, we have the upper bound
    $\Order{R(N+M) + R^2[(n-1)N + (m-1)M]}.$

    The result for \textsc{mpo} is found in a similar way using the fact that now we have only one set of indices
$(h_1, \ldots, h_n)= ( i_1 + (j_1 -1) I_1, \ldots, i_n + (j_n -1) I_n )$
of dimension $N M$.\qedsymbol

An important corollary of the above result is the comparison at inference time, that is, in the computation of the
forward pass, between \textsc{mps} and \textsc{mpo}, and the full model multiplication $\bW \bx$.

\begin{corollary}\label{prop:2}
Let $\kappa^{-1} > 1$ be a given compression rate. Suppose that $n=m$ in the factorization of the dimensions of $\bW$.
For large $N$ and/or $M$, 
  $\nicefrac{O_{\textsc{mps}}}{NM} = \kappa [ \nicefrac{(N+M)}{(I+J)}]$, and
  $\nicefrac{O_{\textsc{mpo}}}{N M} = \kappa [\nicefrac{N M}{IJ}]$.
In particular: 

  \noindent (a) inference with \textsc{mps} is more efficient than reconstructing the weight matrix
  if $\kappa <  \nicefrac{(I+J)}{(N+M)}$;
  
\noindent (b) for $n>2$, the efficiency gain of \textsc{mps} over \textsc{mpo} at inference
  $\nicefrac{O_{\textsc{mpo}}}{ O_{\textsc{mps}}} =  \nicefrac{NM (I+J)}{[IJ (N+M)]}$; for $n=2$,
  $\nicefrac{O_{\textsc{mpo}}}{ O_{\textsc{mps}}}= \nicefrac{NM (I+J)}{[2 IJ (N+M)]}$.
\end{corollary}
  
\paragraph{Proof:}
We suppose that $n=m$. From Theorem~\ref{theo:1}, for a given compression rate $\kappa$ we need to have
$\kappa MN = (I+J)[R_s + (n-1) R_s^2] = I J[ 2R_o + (n-2) R_o^2]$, where $R_s, R_o$ are the corresponding inner ranks of \textsc{mps} and \textsc{mpo}.
This yields $R_s = [2(n-1)]^{-1}( \sqrt{1 + \nicefrac{(4\kappa (n-1) NM)}{(I+J)}\ } - 1)$,
and $R_o= [(n-2)]^{-1}( \sqrt{1 + \nicefrac{(\kappa (n-2)NM)}{(IJ)}\ } - 1)$, for $n>2$, and
$R_o = \kappa NM /(2 IJ)$, for $n=2$.
This implies $O_{\textsc{mps}} = \kappa NM \bigl(\nicefrac{(N+M)}{(I+J)}\bigr)$.
For large $N$ and/or $M$, this also implies
 $O_{\textsc{mpo}} = \kappa \bigl(\nicefrac{N^2M^2}{IJ}\bigr)$, for $n>2$;
for $n=2$, $O_{\textsc{mpo}} = \bigl( \nicefrac{\kappa}{2}\bigr) \bigl( \nicefrac{N^2M^2}{I J}\bigr)$.

In particular, inference with \textsc{mps} is more efficient that reconstructing the weight matrix when
$\kappa < \nicefrac{(I+J)}{(N+M)}$.
This shows part (a) of the statement.
For part (b), we just need to compute
the efficiency gain of \textsc{mps} over \textsc{mpo} which by definition, and given the computations above is easily seen to be
$NM(I+J) /([N+M]IJ)$, for $n> 2$. This ratio is halved when $n=2$.\qedsymbol

\noindent {\bf Remarks:}
The theorem shows that \textsc{mps} and \textsc{mpo} tensor train decompositions are very effective at compressing a weight matrix.
However, the \textsc{mpo} bound is larger than the case where the dimensions are not intertwined (\textsc{mps}).
The key to keep this bound small is to combined the row dimensions $\{I_1, \ldots, I_n\}$
with a permutation of the column dimensions $\{J_{\pi(1)}, \ldots, J_{\pi(n)}\}$
so that  $I_k J_{\pi(k)}$ are kept as small as possible.

    Assuming that $M=N$ and $J=I$,
    the efficiency gain of \textsc{mps} in the multiplication is of order $\Order{N/I}$.
    Assuming that $I= N^{1/n}$, we get an efficiency of order $N^{(n-1)/n}$.
    So there are great gains at  inference time when \textsc{mps} are used.

\section{Initialization and layer normalization for tensor weights}
\label{sec:initialization}
The usual way to initialize the weights is by random assignment (e.g., by Glorot initialization \citep{glorot-bengio-2010}).
This is usually done
with a Uniform$(-\nicefrac{1}{\sqrt{M}},\, \nicefrac{1}{\sqrt{M}})$ or a Normal$(0, \nicefrac{1}{\sqrt{M}})$ \citep{tjandra-et-al-2017}.
When the weights are given by product of tensors, what need to be initialize are the tensor components.
The typical weight is given by the form
$w = \sum_{k_o, k_1,k_2,\ldots, k_{n-1}} a^{(1)}_{k_o,i_1, k_1} a^{(2)}_{k_1, i_2, k_2}\cdots  a^{(n)}_{k_{n-1}, i_n, k_o}.$
So we need to know the
distributions of products and sums of products. This is not straightforward as sum of products of
uniform or normal variables does not result in a uniform or normal variable.
But since usually $n$ is large, and assuming identically distributed terms,
by the central limit theorem, the sum should be closely distributed as a normal variable.
So we just need to know its mean and variance.
Assuming that all individual tensor terms are centered at zero yields zero-mean terms.
For the variance, assuming that
$\operatorname{Var}(a^{(j)}_{k_{j-1},i_j,k_j}) = \sigma^2$,
gives
  $\operatorname{Var}(w) =  \sigma^{2n} \prod_{k=0}^{n-1} r_k.$
To obtain a normally distributed weight with variance $\nicefrac{1}{\sqrt{M}}$ we set
$\sigma^2 = [ \prod_{k=0}^{n-1} r_k ]^{-1/n} \! \! \! \! M^{-1/(2n)}$.
Alternatively, if we would like to have a more uniform distribution, say Uniform$(-B, B)$ for some
$B>0$ (e.g., $B = \nicefrac{1}{\sqrt{M}}$), we ask for the resulting normal distribution
to
be flat in the region $(-B,B)$. That is, we ask
$\operatorname{Prob}( \lvert W\rvert \leq B ) = 1 -\alpha$ for some small $\alpha >0$.
This implies
$\sigma^{2} = [B / \Phi^{-1}(  1 -  \nicefrac{\alpha}{2}) ]^{2/n} [\prod_{k=0}^{n-1} r_k]^{-1/n},$
where $\Phi(\cdot)$ denotes the standard normal cumulative distribution function.
If instead, the individual term distributions are Uniform$(-b,b)$, we obtain
$b = \sqrt{3} [B/\Phi^{-1}(  1 - \nicefrac{\alpha}{2} ) ]^{1/n} [\prod_{k=0}^{n-1} r_k ]^{-1/(2n)}.$

\paragraph{Normalization.}
Without any constraints on the core tensors, the tensor decomposition is not necessarily unique \citep{kolda-bader-2009}.
Also, because the weights are computed as products of core tensors, backpropagation computations for a given
tensor weight involves the weights of all the other tensors.
These two facts make training of tensor-compressed RNN harder than training of non-constrained RNN.  
In our experiments, layer normalization \citep{ba2016layer} stabilized the tensor weights. So we have adopted
this technique as standard when training tensor-compressed RNN.

\section{Knowledge distillation}
\label{sec:KD}
It is harder for a compressed network to achieve the same level of accuracy than a larger network.
Instead of training a compressed network from scratch, one can use the knowledge gained by a larger network
to help training the compressed network. Knowledge distillation \citep{Hinton-2014,Howard-et-al-2017, Polino-et-al-2018,Phuong-Lampert-2019} is a technique widely used to mitigate the problem.
The classification scores from the larger network are incorporated in the
loss function associated with the training of the smallest network.
This is usually done through a cross-entropy loss.
For a linear output network (that is, with no sigmoid transformation of the output), the ``cross-entropy'' is replaced by
squared errors on the activations. That is, if $\bW^* \bx$ and $\bW\bx$ are, respectively, the linear activations of a given input vector $\bx$
for the larger and smaller networks, then {\it knowledge distillation on activations}  corresponds
to adding to the loss function the sum of squared errors $\sum_i\lVert \bW^* \bx_i - \bW \bx_i\rVert^2_2$, where
$\lVert \cdot \rVert_2$ denotes the Euclidean distance. This is the same as
$\operatorname{Trace}\bigl[ (\bW^* -\bW) \bS (\bW^* -\bW)^t\bigr]$, where $\bS = \sum_i \bx_i \bx_i^t$ is the covariance matrix
of the data (assuming that the input vectors are centered), and the superscript $\cdot^t$ indicates matrix transposition.
The importance of the structure in $\bW^*$ depends on the data weights $\bS$. The smaller compressed network
might perform as well as the larger one if $\bS$ puts very little weight on a large number of elements of $\bW^* -\bW$. Since $\bS$
is symmetric semi-positive definite, these weights will depend mostly on the eigenvalues of $\bS$. A large difference between the largest and smallest
eigenvalues of $\bS$ should indicate that compression is feasible.
If $\bS$ is replaced by the identity matrix we obtain {\it knowledge distillation on weights} since the added loss
$\operatorname{Trace}\bigl[ (\bW^* -\bW) (\bW^* -\bW)^t\bigr] = \lVert \bW^* - \bW\rVert_2^2$ becomes the squared of the Frobenius norm of
the difference between the weights. One can also see this last setup as a regularization or prior knowledge
on the tensor decomposition.
In our experiments we have tried these two linear distillation techniques. The original loss function $L(\bW)$ is replaced by
${\cal L}(\bW, \lambda) = L(\bW) + \lambda \operatorname{Trace}\bigl[ (\bW^* -\bW) \bS (\bW^* -\bW)^t\bigr]$, for some appropriate
value of the hyper-parameter $\lambda$.
Knowledge distillation is done by first training a full non-compressed network, and then using the last value of $\bW^*$ to start training
a tensor-compressed network.

\section{Experiments}
\label{sec:experiments}
For our experiments,
we set up a language model task.
This consists of estimating probabilistic sentence models $p(w_1)p(w_2|w_1)p(w_3|w_2w_1) \cdots p(w_T|w_{T-1}\cdots w_1)$, for each string of words
$w_1\ldots w_T.$
Each \textsc{lstm} cell in the network models one of these conditional probabilities.
Our architecture may be seen in Figure~\ref{fig:ourLSTM}.
\begin{figure}[ht!]
  \centering
  \rule[-.5cm]{0cm}{2.5cm}
  \includegraphics[width=0.6\linewidth]{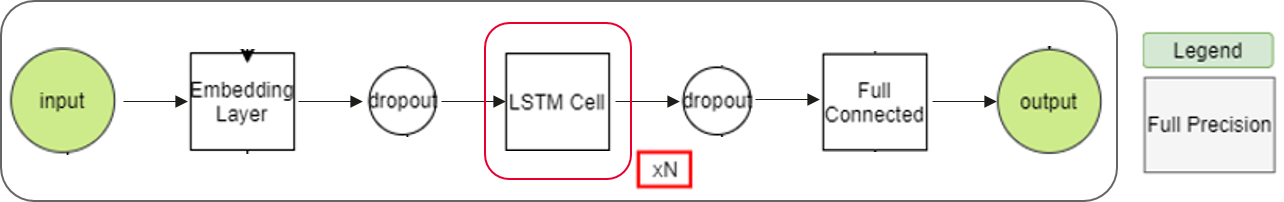}
  \rule[-.5cm]{0cm}{2cm}
  \caption{The \textsc{lstm} network architecture for the Penn Treebank and the Google One-billion dataset}
  \label{fig:ourLSTM}
\end{figure}
The task was run on two largely different size corpuses, both publicly available:
The Penn Treebank (\textsc{ptb}) dataset \citep{taylor2003penn}, and the \textit{Google One Billion Words} (Google corpus) dataset \cite{chelba2013billion}.
Both corpuses are widely used in machine learning and \textsc{nlp} research \citep{devlin2018bert,NIPS2017_7159}.
We recall that a treebank is a parsed text corpus with annotated semantic and/or syntactic structure. 

\textsc{ptb}, which was built in the period 1989-1996, is one of the first large such corpus.
It comprises millions of words from annotated text from diverse sources such as the Wall Street journal.
For our experiments, we used the built-in Pytorch Penn Treebank corpus \citep{Pytorch-PTB}.
The training and test sets comprise, respectively, 929,580 and 82,420 sentences.
The vocabulary size is 10,000 words.

The Google corpus is a much larger corpus  and  a harder task than \textsc{ptb}.
It was built from
the 2011  Workshop on Statistical Machine Learning
\footnotemark\footnotetext{https://www.statmt.org/wmt11/translation-task.html}
by only selecting the English corpora.
It has a vocabulary size of 793,471 words.
For our experiments, we selected the first 10 million sentences of the corpus, and gather
the 100,000 most frequent words to form our vocabulary.
The training set consists of 7 millions sentences. The test set contains 2 millions sentences.
We measure the performance of the compressed models with {\it perplexity.} This is a goodness-of-fit
measure commonly used in \textsc{nlp} tasks. Its logarithm is proportional to the
log-likelihood.
The comparisons of perplexity are made on the testing subsets of the corpuses.

\subsection{Perplexity versus compression}
Our first experiment consists of measuring the performance
of \textsc{mps}-based and \textsc{mpo}-based tensorized \textsc{lstm} architectures
under different compression rates ranging from 1.59 to more than 53.
For this task we choose the \textsc{ptb} corpus. \citep{taylor2003penn}.

We consider three different decompositions for the weight matrices, yielding
two, three or four cores for \textsc{mpo}, and four, six and eight cores for \textsc{mps}.
The compression rates correspond to different values of the inner rank parameters.
For this experiment we fix the inner rank parameters to the same value $R$.
For each type of decomposition \textsc{mps} and \textsc{mpo} we try different values of $R$ ranging from $20$ to $365$,
so as to obtain similar compression rates for both models.
In the results below, the rates has been binned to ease their comparison.
The \textsc{mps} and \textsc{mpo} architectures compared are described in Table~\ref{table:tensor:train:architectures}.
\begin{table}[ht!]
  \caption{Tensor train architectures considered for the Penn Treebank and Google corpuses}
  \label{table:tensor:train:architectures}
  \centering
\scalebox{0.80}{  \begin{tabular}{ccccc}
    \toprule
             & Number of  &\multicolumn{2}{c}{\textsc{mps}} & \textsc{mpo} \\
    Corpus   &  factors   & $(I_1, \ldots, I_n)$ & $(J_1,\ldots,J_m)$   &  $(I_1 J_1, \ldots, I_n J_n)$ \\
    \midrule
Penn            & 2          & $(50, 52) $     & $(25,26)$    & $(1250, 1352)$ \\
Treebank        & 3          & $(13, 10,20) $  & $(13,5, 10)$ & $(169, 50, 200)$ \\
                & 4          & $(10,5,4, 13) $ & $(5,5,13,2)$ & $(50, 25,52,26)$ \\ 
    \midrule
 Google 1 Billion           & 2          & $(64, 128)$ & $(16,128)$ & $(1250, 1352)$ \\
\bottomrule
\end{tabular}}

\end{table}
In our experiments, we set the sentence length to $T=35$ words.
The embedding size or dimension of the word vector was set to 650.
The hidden dimension of the \textsc{lstm} cell was also fixed at 650. We also used a batch of size 20.
There are four weight matrices associated with the input vector $\bx$, and four weight matrices
associated with the \textsc{lstm} hidden output vector $\bh$. By stacking the weight
matrices together, we form two matrices $\bW_{x}$ and $\bW_{h}$
of sizes $( 4\times 650) \times 650$ associated with $\bx$ and $\bh$, respectively.
Each one of these matrices was represented as a tensor train, either \textsc{mps} or \textsc{mpo}, from the
start of the training of the models. That is, in this task, the models considered are fully
tensorized from the beginning. The exception is the {\it full model}, which is 
the model without any tensor decomposition on the weight matrices.
The compression rates are computed as
number of weight parameters of the full model divided by the number of weight parameters of the tensorized model
given in Theorem~\ref{theo:1}.

For \textsc{ptb}, the full model perplexity is $79.91$. It contains $3.38\times 10^6$ parameters.
The perplexity results associated with five compression rate bins for the different tensorized \textsc{lstm} models
are shown in Table~\ref{table:tensor:train:ptb}, and also in the leftmost panels of Figure~\ref{fig:ptb:perplexity}.
Together with the main statistics associated with the compression rate bins, we also show the minimum
perplexity achieved in each compression bin.

\begin{table}[ht!]
  \caption{\textsc{mps} and \textsc{mpo} tensor train perplexity results on the Penn Treebank corpus.
  \textsc{sd} stands for standard deviation; ``at rate'' is the compression rate associated with the minimum perplexity
  at the corresponding compression bin. Numbers in boldface are the minimum perplexities achieved.}
  \label{table:tensor:train:ptb}
  \centering
\scalebox{0.90}{  \begin{tabular}{ccccc}
    \toprule
       \mbox{\  }   & \multicolumn{4}{c}{Tensor Train Decomposition}  \\
  \multicolumn{1}{c}{Compression}    &  \multicolumn{2}{c}{\textsc{mps}} & \multicolumn{2}{c}{\textsc{mpo}} \\
  Rate           &  Mean (\textsc{sd}) & Minimum [at rate] & Mean (\textsc{sd}) & Minimum [at rate] \\
  \midrule
  1 - 1.8   &  $83.11\ (0.47)$ &  $82.78\ [1.51]$ & $83.13\ (0.00)$ & $83.13\ [1.51]$ \\
  1.8 - 2.6 &  $84.73\ (3.11)$ &  $\mathbf{82.03\ [1.79]}$ & $83.39\ (0.79)$ & $82.50\ [1.79]$ \\
  2.6 - 3.4 &  $85.88\ (3.43)$ &  $82.94\ [2.68]$ & $83.76\ (1.04)$ & $\mathbf{82.07\ [2.68]}$ \\
  3.4 - 6   &  $86.89\ (3.22)$ &  $83.13\ [3.38]$ & $85.04\ (1.04)$ & $83.46\ [3.82]$ \\
  6 - 53    &  $87.68\ (3.34)$ &  $83.91\ [6.00]$ & $89.06\ (3.16)$ & $85.30\ [6.49]$ \\
  \bottomrule
  \end{tabular}}
\end{table}

\begin{figure}[ht!]
    \centering
  \rule[-.5cm]{0cm}{2.5cm}
    \scalebox{0.80}{\includegraphics[width=0.8\textwidth]{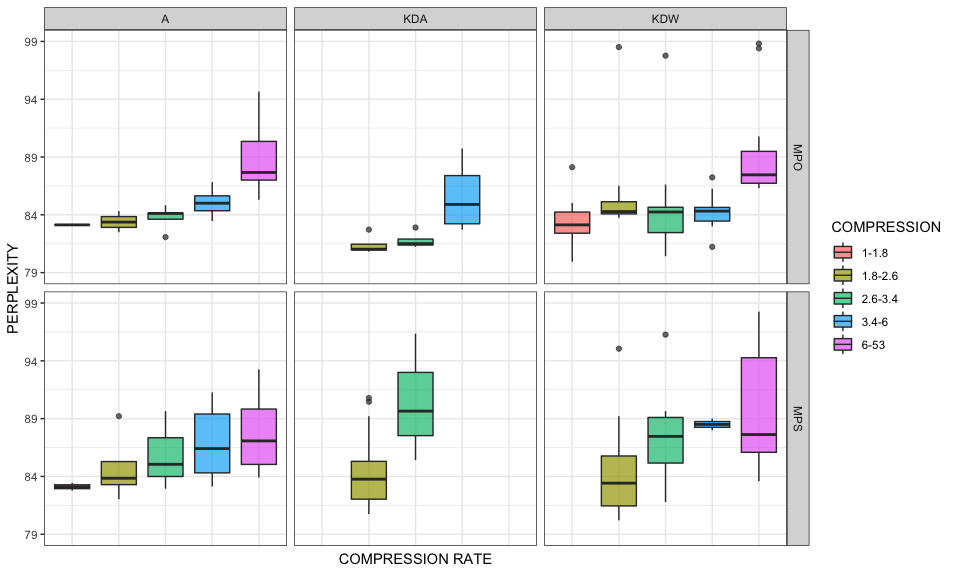}}
   \caption{Perplexity of different tensor train models on the Penn Treebank corpus.
     \textsc{kdw} stands for knowledge distillation on weights, \textsc{kda} for knowledge distillation on activations,
     and A for {\it alone,} that is, no knowledge distillation. }
   \label{fig:ptb:perplexity}
\end{figure}

\subsection{Performance with knowledge distillation}
\label{sec:ptb:KD}
As expected, the best results are linked to lower compression rates.
In fact, perplexity increases nearly quadratically with compression, specially for \textsc{mpo} models
(see panel A of  Figure~\ref{fig:ptb:perplexity}).
Our second experiment is to study the improvement, if any, associated with using knowledge
distillation in the training of tensorized models. For this task, we only look at the perplexity results
of the lower compression models two-core \textsc{mpo} and four-core \textsc{mps}. The results with knowledge distillation on activations (\textsc{kda})
and on weights (\textsc{kdw}) are displayed in the middle and rightmost panels of Figure~\ref{fig:ptb:perplexity}.
The statistics are also displayed in Table\ref{table:ptb:KD}.
The perplexity means and standard deviations are computed from
different models with compression rates in the corresponding bins, as well as, from different values of
the hyper-parameter $\lambda \in\{0.5, 1, 5, 10, 50, 100, 500, 5000\}\times 10^{-6}$ (see Section~\ref{sec:KD}).
\begin{table}[ht!]
  \caption{Knowledge distillation on \textsc{mps} and \textsc{mpo} tensor trains. Perplexity results on the Penn Treebank corpus. Numbers in boldface are the minimum perplexities achieved.
  }
  \label{table:ptb:KD}
  \centering
\scalebox{0.90}{\begin{tabular}{c|ccccc}
    \toprule
 &       \mbox{\  }   & \multicolumn{4}{c}{Tensor Train Decomposition}  \\
 &  \multicolumn{1}{c}{Compression}    &  \multicolumn{2}{c}{\textsc{mps}} & \multicolumn{2}{c}{\textsc{mpo}} \\
 &  Rate           &  Mean (\textsc{sd}) & Minimum [at rate] & Mean (\textsc{sd}) & Minimum [at rate] \\
  \cline{2-6}
        &  1 - 1.8   &      -           &  -               & $83.40\ (1.91)$ & $\mathbf{79.95\ [1.78]}$ \\        
\textsc{kd} &  1.8 - 2.6 &  $84.17\ (3.73)$ &  $\mathbf{80.20\ [1.80]}$ & $86.07\ (4.74)$ & $83.74\ [1.80]$ \\
 on     &  2.6 - 3.4 &  $87.83\ (4.96)$ &  $81.79\ [2.68]$ & $84.72\ (4.08)$ & $80.42\ [2.68]$ \\
weights &  3.4 - 6   &  $88.50\ (0.71)$ &  $88.00\ [3.74]$ & $84.25\ (1.47)$ & $81.23\ [3.82]$ \\
        &  6 - 53    &  $89.86\ (5.94)$ &  $83.59\ [6.99]$ & $89.29\ (4.33)$ & $86.03\ [6.00]$ \\
\midrule
& & & & & \\
\textsc{kd} &   1.8 - 2.6 &  $84.34\ (3.02)$ &  $\mathbf{80.74\ [1.80]}$ & $81.40\ (0.89)$ & $\mathbf{80.82\ [1.80]}$ \\
 on         &   2.6 - 3.4 &  $90.47\ (5.52)$ &  $85.41\ [2.60]$ & $81.79\ (0.76)$ & $81.23\ [2.68]$ \\
activations &   3.4 - 6   &  -               &  -               & $85.58\ (2.64)$ & $82.72\ [3.82]$ \\
  \bottomrule
  \end{tabular}}
\end{table}
The results are similar for \textsc{mps} and \textsc{mpo}, achieving perplexities very close to the full model
for the lowest compression rates near 1.8. We stress that this compression rate corresponds to
a reduction of 44\% in the number of weight parameters.

\subsection{Performance of tensorized \textsc{lstm} cells with knowledge distillation for the Google corpus}
Our third experiment consists of trying tensor train compression on
a large dataset.
Due to time and computer memory constraints,
we only run our experiments with four-core \textsc{mps} (\textsc{mps}2) and two-core \textsc{mpo} (\textsc{mpo}2)
with knowledge distillation. This choice is based on the good performance
of this setup for the smaller Penn Treebank corpus.
The tensor-train setup is summarized in Table~\ref{table:tensor:train:architectures}.
For this experiment, we set the sentence length to $T=20$ words.
The embedding size or dimension of the word vector was set to 256.
The size of the \textsc{lstm} cell was fixed to 2048. 
The full model contains $18.8\times 10^6$ parameters.

For the subset of the Google corpus used in this experiment, the full model perplexity is $67.48$.
Hence, the task appears simpler than for the \textsc{ptb} corpus, which had a larger perplexity.
But in reality, the task is harder due to the vocabulary size. The size of the model and the amount of
training data might explain the difference in perplexities.
A summary of the results is displayed in Table~\ref{table:google}.
The perplexity means and standard deviations are computed as explained in Section~\ref{sec:ptb:KD}.
The \textsc{mpo} tensorized \textsc{lstm} architecture reaches perplexities better than the full model.
The \textsc{mps} architecture yield perplexities close to but slightly larger than the full model.

\begin{table}[ht!]
  \caption{Knowledge distillation on \textsc{mps} and \textsc{mpo} tensor trains. Perplexity results on the Google corpus.
    \textsc{sd} stands for standard deviation. Numbers in boldface are the minimum perplexities achieved.}
  \label{table:google}
  \centering
\scalebox{0.90}{\begin{tabular}{c|ccccc}
    \toprule
Tensor &              & \multicolumn{4}{c}{Knowledge Distillation}  \\
 Train & Compression  &  \multicolumn{2}{c}{\textsc{kdw}} & \multicolumn{2}{c}{\textsc{kda}} \\
       &  Rate        &  Mean (\textsc{sd}) & Minimum & Mean (\textsc{sd}) & Minimum \\
  \midrule
\textsc{mpo}  &  2.5   &  $70.59\ (2.19)$ & $\mathbf{69.04}$ & $65.95\ (0.06)$  & $\mathbf{65.9}$ \\
     &  6.7   &  $70.31\ (1.59)$ & $69.35$ & $67.55\ (0.63)$ & $67.1$ \\
\midrule
& & & & & \\
     &   2.5 &  $71.60\ (2.59)$ &  $\mathbf{69.45}$ & $71.94\ (2.64)$ & $69.62$ \\
\textsc{mps}  &   5.6 &  $73.56\ (2.41)$ &  $71.69$ & $72.44\ (2.76)$ & $\mathbf{69.4}$ \\
     &   7.7 &  $73.67\ (2.14)$ &  $71.9$  & $73.76\ (1.97)$ & $71.33$ \\
  \bottomrule
  \end{tabular}}
\end{table}

\subsection{Benchmarking computing time at inference}
\label{sec:benchmark}
As stressed in the introduction and in Section~\ref{sec:storage:time}, \textsc{mps} \textsc{lstm} cells are much more efficient in
computing the product $\bW\bx$. In this experiment we compare the times needed to do this computation
for \textsc{mps} and \textsc{mpo} tensorized networks. We benchmarked the inference times (i.e., forward pass times)
for two, three and four-core \textsc{mpo}, as well as for four, six and eight-core
\textsc{mps} tensorized \textsc{lstm}s on \textsc{cpu}, at a fixed compression rate of about 1.8.
The time was inferred using 1327 sentences of the Penn Treebank corpus.
We run the inference task twelve times. The first two measures were discarded so as to get rid
of hardware-related issues at the start of the experiment (i.e., loading in the cache, etc.).
Table~\ref{table:benchmark} shows the results in real seconds.
The experiments were run on a 20-core
Intel Xeon Processor E5 2698v4 \textsc{cpu} with 256 \textsc{gb} of memory \textsc{ram}.
Although, the multiplication after reconstruction has been optimized for \textsc{mpo}, \textsc{mps} is still faster.
\begin{table}[ht!]
  \caption{Mean inference times for \textsc{mps} and \textsc{mpo} tensor trains on the Penn Treebank corpus.
    The corresponding standard deviations are shown in parentheses.}
  \label{table:benchmark}
  \centering
\scalebox{0.90}{\begin{tabular}{cccccc}
    \toprule
    \multicolumn{6}{c}{Tensor Train Decomposition} \\
    4-core \textsc{mps} & 2-core \textsc{mpo} & six-core \textsc{mps} & 3-core \textsc{mpo} & 8-core \textsc{mps} & 4-core \textsc{mpo} \\ \midrule
    $1.39\ (0.08)$ & $3.64\ (0.12)$ & $1.90\ (0.05)$  &  $4.25\ (0.13)$ & $2.40\ (0.07)$ & $6.77\ (0.23)$ \\
    \bottomrule
    \end{tabular}}
\end{table}

\section{Conclusions}

Our goal is to find small models that perform well, as opposed to models that are just compressed networks for inference.
Tensor decompositions greatly reduce the amount of storage both during training and inference.
They also have the potential of accelerating
computations at inference time in small storage and compute constraint devices.
Our experimental results show that \textsc{lstm} networks can be effectively compressed via tensor decompositions with
some gain or very little lost in performance.
Short \textsc{mps} and \textsc{mpo} tensor trains, i.e., tensor decompositions with very few large inner rank tensors,
appear to perform better.
We show that up to 6x compression rates may be achieved without hurting by much perplexity.
However, compression rates of about 2x seem optimal.
Our results applied to both small and very large datasets.
In our experiments, the compressed \textsc{lstm} networks trained with knowledge distillation performed better on the large dataset.
We believe that the reasons  for this are the following:
(a) the training set is very large, thus facilitating parameter
estimation, and more importantly generalization; (b) the covariance matrix of the data $\bS$ must have several
small eigenvalues which allows a more effective knowledge distillation on tensor train activations.

As mentioned earlier, compressing with tensors may also introduced computational gains at inference time
if inference relies on \textsc{cpu}s as opposed to \textsc{gpu}s.
This is the current situation on small constraint devices.
At similar compression rates,
tensor decompositions based on \textsc{mpo} appear to perform slightly better than tensor decompositions based on \textsc{mps}.
However, gains at inference time could be obtained only if one does not need to reconstruct the weight matrices of the \textsc{lstm} cell.
Following our arguments of Section~\ref{sec:storage:time}, \textsc{mps} tensor trains present such a property.
Computation as well as compression goals should be balanced in computationally constrained devices.
Our experiments show that \textsc{mps} tensor train decompositions are the right answer to balance these two goals.

\bibliographystyle{plainnat}
\bibliography{referencesNN}

\end{document}